\documentclass[journal]{IEEEtran}
\usepackage[T1]{fontenc}
\usepackage{graphicx}
\usepackage{times}
\usepackage{helvet}
\usepackage{courier}
\usepackage{amsmath}
\usepackage{algorithm}
\usepackage{algpseudocode}
\usepackage{csquotes}
\usepackage{paralist}
\usepackage{amssymb} 
\usepackage{indentfirst}
\usepackage{subfigure}
\usepackage{float}
\usepackage{multirow}
\usepackage{cite}
\usepackage{bm}
\usepackage{mathrsfs}

\usepackage{amsfonts}
\usepackage{todonotes}
\usepackage{pgfplots}  
\usepackage{epstopdf}
\usepackage{epsfig}
\pgfplotsset{compat=newest}
\usepackage{color}
\usepackage{xcolor}
\usepackage[colorlinks, linkcolor=red,  anchorcolor=blue, citecolor=citecolor]{hyperref}
\definecolor{forestgreen}{RGB}{0,139,69}
\definecolor{citecolor}{HTML}{0071bc}
\definecolor{SeaGreen4}{RGB}{0,205,102}
\definecolor{SlateBlue}{RGB}{106,90,205}
\definecolor{DarkRed}{RGB}{178,34,34}
\usepackage[switch]{lineno}
\usepackage{textcomp,booktabs}
\usepackage{pifont} 
\usepackage{makecell}
\usepackage{colortbl}
\definecolor{mygray}{gray}{.9}
\definecolor{mypink}{rgb}{.99,.91,.95}
\definecolor{mycyan}{cmyk}{.3,0,0,0}

\begin{document}

\title{Active Adversarial Perturbation-driven Associative Memory Retrieval for RGB-Event Visual Object Tracking}

\author{Xiao Wang, \emph{Member, IEEE}, Xufeng Lou, Zikang Yan, Lan Chen*, Sibao Chen, 
        Yaowei Wang, \emph{Member, IEEE}, \\ Yonghong Tian, \emph{Fellow, IEEE}, Jin Tang

\thanks{$\bullet$ Xiao Wang, Xufeng Lou, Zikang Yan, Sibao Chen, and Jin Tang are with the School of Computer Science and Technology, Anhui University, Hefei 230601, China. (email: \{xiaowang, sbchen, tangjin\}@ahu.edu.cn, louxufeng@stu.ahu.edu.cn, e24201056@stu.ahu.edu.cn)} 

\thanks{$\bullet$ Lan Chen is with the School of Electronic and Information Engineering, Anhui University, Hefei 230601, China. (email: chenlan@ahu.edu.cn)} 

\thanks{$\bullet$ Yaowei Wang is with Peng Cheng Laboratory, Shenzhen, China; Harbin Institute of Technology, Shenzhen, China (email: wangyw@pcl.ac.cn)} 

\thanks{$\bullet$ Yonghong Tian is with Peng Cheng Laboratory, Shenzhen, China; School of Computer Science, Peking University, China; School of Electronic and Computer Engineering, Shenzhen Graduate School, Peking University, China (email: yhtian@pku.edu.cn)}

\thanks{* Corresponding Author: Lan Chen}  
}

\markboth{ IEEE Transactions on ***, 2026 }
{Shell \MakeLowercase{\textit{et al.}}: Bare Demo of IEEEtran.cls for IEEE Journals}

\maketitle

\begin{abstract}
RGB-Event tracking improves localization robustness by fusing RGB appearance textures and dense temporal motion cues from event sensors. While this multi-modal scheme broadens tracking applicability, real-world scenes suffer diverse structured signal degradations that hinder traditional multi-modal fusion. In harsh environments, either modality can lose reliability drastically, and targets frequently appear incomplete due to occlusion, edge truncation and foreground clutter.
To tackle the above challenges, we present a hierarchical perturbation and retrieval framework tailored for RGB-Event tracking with robustness against partial target missing and modal degradation, termed APRTrack. To mimic real-world signal corruption, APRTrack constructs structured degradation via two adversarial perturbation branches at the modality and spatial levels, which separately simulate full-modal failure and localized target region absence. A hierarchical routing mechanism is designed to disentangle the training pipelines of the two perturbation types, effectively eliminating feature collapse induced by superimposed degradation constraints. Furthermore, we devise Footprint-guided Channel-calibrated Hopfield Retrieval (FCHR) for reliable historical information compensation. This module evaluates retrieval confidence based on association footprints between queries and memory banks, and calibrates the retrieval metric space prior to Hopfield matching, realizing controllable historical feature compensation bounded to target regions. 
Extensive experiments on FE108, COESOT, VisEvent, and FELT datasets demonstrate the effectiveness of our proposed strategies for the RGB-Event visual object tracking. The source code and pre-trained models will be released on \url{https://github.com/Event-AHU/OpenEvTracking}.
\end{abstract}


\IEEEpeerreviewmaketitle

\section{Introduction}

\begin{figure}
\centering
\includegraphics[width=0.9\linewidth]{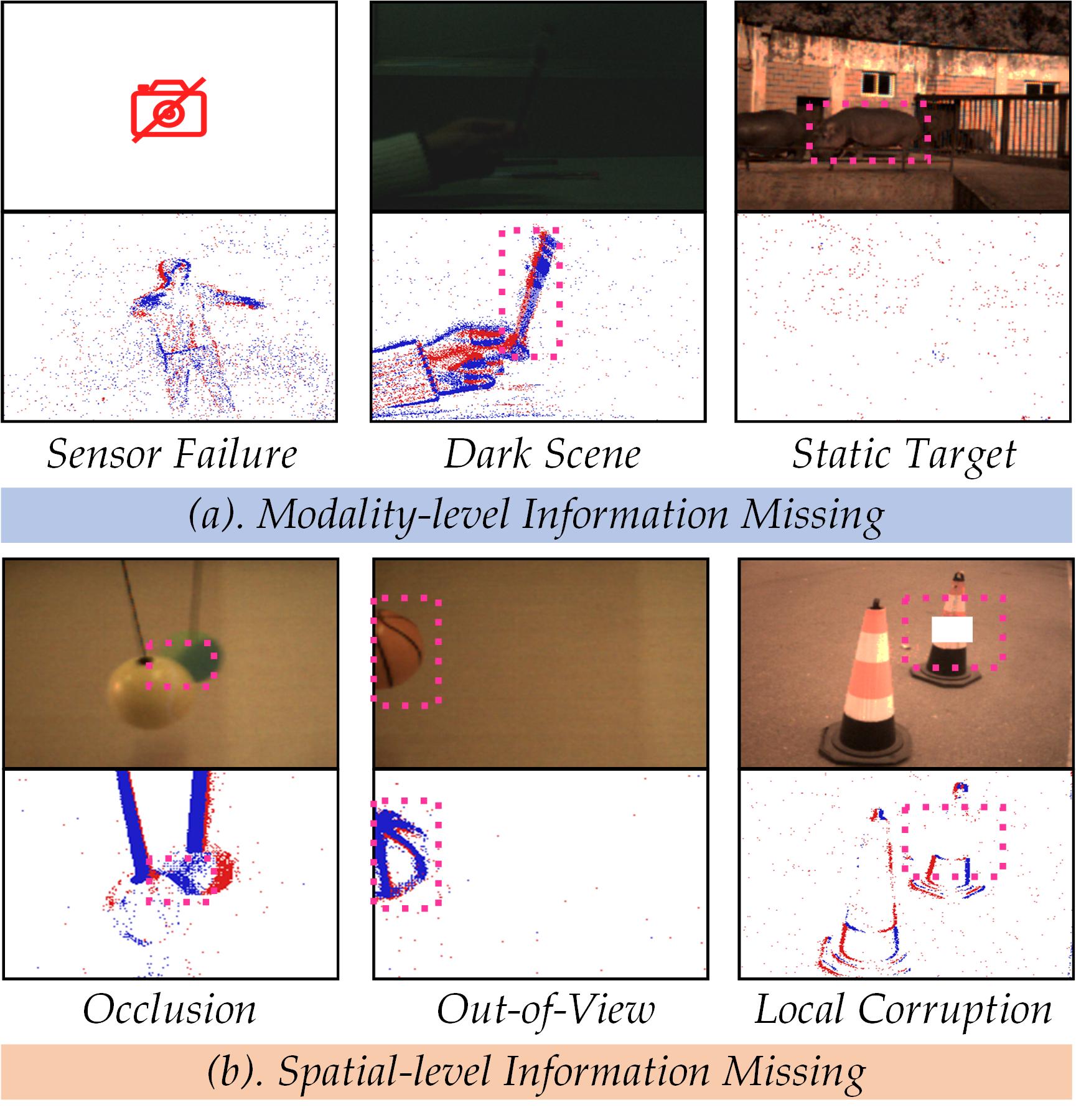}
\caption{Representative modality-level and spatial-level missing issues in RGB-Event tracking.} 
\label{fig:intro}
\end{figure}

\IEEEPARstart{S}{ingle} Object Tracking (SOT) aims to predict the bounding box of a specified target in subsequent video frames given its initial annotated state. As a foundational research task in computer vision, SOT has been widely deployed in intelligent surveillance, autonomous driving, robotic perception, and unmanned intelligent systems. Benefiting from the booming development of deep learning, modern deep trackers, particularly Transformer-based tracking architectures~\cite{chen2021transformer, cui2022mixformer, ye2022joint, chen2023seqtrack}, have achieved advanced target feature representation and long-range temporal dependency modeling, yielding promising performance against common challenges including scale variation, arbitrary pose transformation, and complex background clutter. Nevertheless, pure RGB-based tracking still suffers from inherent limitations in harsh imaging conditions: low ambient illumination, camera overexposure, severe motion blur, and full/partial occlusion will greatly degrade target appearance discriminability and impair long-term localization stability.

To mitigate the above drawbacks, event cameras have emerged as a complementary bio-inspired sensing modality for visual tracking. Different from standard frame-based cameras that capture integral RGB frames at fixed sampling rates, event cameras fire asynchronous pixel-level events triggered by logarithmic luminance changes, featuring ultra-high temporal resolution, wide dynamic range, and low redundant data output~\cite{gallego2020event, huang2018event}. Such asynchronous event streams can capture fine-grained motion dynamics that are invisible for RGB frames under fast camera movement, drastic illumination fluctuation and dim light scenarios. However, event data has inherent defects as well: event generation relies heavily on luminance variation, which provides extremely sparse semantic information for static or slowly moving targets, and lacks color perception and detailed textural features. Given the inherent complementary characteristics between RGB and event modalities, RGB-Event cross-modal collaborative learning has evolved into a mainstream research direction for robust visual tracking~\cite{zhang2021object, wang2023visevent, tang2025revisiting, wang2024long}. 

Existing RGB-Event trackers mainly focus on cross-modal feature fusion, unified representation learning, and temporal cue modeling~\cite{zhang2023frame, zhu2023cross, zhu2023visual, fu2023distractor, zhang2024revisiting, tang2025revisiting, wang2024long, sun2025exploring}, aiming to better exploit the complementarity between RGB and Event modalities. These studies have shown the potential of joint RGB-Event modeling and moved the field from simple modality concatenation toward finer interaction. However, most of them still assume intact inputs and pay limited attention to the structure of input degradation. In real scenarios, modality-level missing can arise from two common sources: direct sensor failure or severe modality degradation that leaves one stream largely uninformative. For example, RGB observations may fail over large regions under low light or overexposure, while event streams may provide weak or noisy responses under low motion or heavy noise. Meanwhile, the target may be only locally missing because of occlusion or truncation, rather than entirely unavailable. These cases correspond to modality-level missing and local target missing, and require different robustness mechanisms, as shown in Fig.~\ref{fig:intro}.

To address the aforementioned issues, this paper proposes APRTrack, an active adversarial hierarchical perturbation and associative memory retrieval framework for missing-robust RGB-Event visual object tracking. APRTrack decomposes real-world structured degradations into hierarchical training constraints, generates hard target-missing samples via adversarial perturbation sampling, and adopts regulated associative memory compensation to recover missing target features. Specifically, APRTrack builds dual perturbation branches after patch embedding: \textit{modality-level perturbation} simulates full failure of RGB or Event modality, while \textit{spatial-level perturbation} models local occlusion and structural target absence. 
The former applies adversarial mutually exclusive modality gating to select challenging modal missing cases, enabling the tracker to localize targets with only one available sensor stream; the latter utilizes adversarial spatial scoring to sample target-limited continuous occlusion regions, ensuring simulated degradation fits real occlusion patterns rather than random token dropout. 
A hierarchical routing strategy is proposed to decouple the training pipelines of the two perturbations, avoiding feature collapse caused by combined modal failure and spatial missing. Besides, we design Footprint-guided Channel-calibrated Hopfield Retrieval (FCHR) for cross-modal historical compensation. Guided by target soft masks, FCHR stores historical target features in memory banks and obtains complementary features via Hopfield associative matching~\cite{ramsauer2020hopfield}. 
To avoid false memory matching, FCHR conducts constrained retrieval: it estimates retrieval reliability from query-memory association footprints, which record how the current query distributes its matches over historical memory, generates channel calibration weights to optimize query metric space, and retrieves target-biased historical features fused into current features as gated residuals.
Accordingly, historical features act as reliability-aware compensation cues, rather than directly replacing original observations. An overview of our proposed APRTrack can be found in Fig.~\ref{fig:framework}. 

To sum up, the main contributions of this work can be summarized as follows:

$\bullet$ We propose a modality-missing-aware RGB-Event object tracking framework named APRTrack. To the best of our knowledge, this is the first tracking approach to explore the effects of multi-level modality missing on multi-modal fusion tracking algorithms, including spatial-level and modality-level missing. 

$\bullet$ We develop an associative memory enhancement module driven by active adversarial perturbations. It can realistically reproduce modality loss and spatial layout information loss occurring in real-world scenes and enhance the robustness and accuracy significantly.  

$\bullet$ Extensive experiments on four large-scale benchmark tracking datasets, including {FE108}~\cite{zhang2021object}, {COESOT}~\cite{tang2025revisiting}, {VisEvent}~\cite{wang2023visevent}, and {FELT}~\cite{wang2024long}, demonstrate the effectiveness of our proposed tracker.

\section{Related Works} 
In this section, we review related works on RGB-Event Object tracking, Missing Modality Learning, and Memory-based Retrieval Networks. More details can be found in the following surveys~\cite{gallego2020event} and paper list~\footnote{\url{https://github.com/wangxiao5791509/Single_Object_Tracking_Paper_List}}.

\subsection{RGB-Event Object Tracking}
RGB-Event tracking benefits from the complementary sensing properties of frame and event cameras. Existing studies have advanced this field from early task formulation and benchmark construction, including FE108~\cite{zhang2021object}, VisEvent~\cite{wang2023visevent}, COESOT~\cite{tang2025revisiting}, FELT~\cite{wang2024long} and high-resolution EventVOT~\cite{wang2024event}, to more effective cross-modal representation learning. Typical designs include frame-event alignment and fusion~\cite{zhang2023frame}, unified Transformer modeling~\cite{tang2025revisiting}, high-rank cross-modal interaction~\cite{zhu2023cross}, prompt-based multi-modal adaptation~\cite{zhu2023visual}, and event modeling for distractor suppression and motion perception~\cite{fu2023distractor,zhang2024revisiting}. Beyond task-specific RGB-Event trackers, unified RGB-X tracking frameworks have also been explored, including UnTrack~\cite{wu2024single}, SUTrack~\cite{Chen2024SuTrack}, SDSTrack~\cite{hou2024sdstrack}, and XTrack~\cite{Tan2025bXTrack}. Recent works further exploit long-term temporal cues, such as AMTTrack~\cite{wang2024long} with the FELT benchmark, MamTrack~\cite{sun2025exploring} with Mamba-based historical modeling, and Mamba-FETrack V2~\cite{wang2025mambafetrack} with state space modeling. However, these methods mainly focus on exploiting available multimodal cues, while the structured missing patterns in RGB-Event tracking remain less explored. Different from these works, APRTrack targets structured missing robustness by combining hierarchical perturbation with historical compensation based on associative memory retrieval.

\subsection{Missing Modality Learning}
Missing modality learning aims to maintain robust multimodal inference when part of the input modalities is unavailable, unreliable, or corrupted by sensor failure. Early studies such as MMIN~\cite{zhao2021missing} and SMIL~\cite{ma2021smil} address uncertain or severely missing modalities by reconstructing missing cues or learning shared multimodal representations. Recent methods further explore different forms of modality recovery and adaptation, including prompt learning for missing modalities~\cite{lee2023multimodal}, shared and specific feature modeling~\cite{wang2023multi}, coherent affective pattern recovery for multimodal sentiment analysis~\cite{huang2026recovering}, data level modality completion with retrieval augmented generation~\cite{he2026rag4dmc}, low rank adaptation for visual recognition with missing modalities~\cite{zhao2026mora}, robust multi modality ReID under missing inputs~\cite{xi2025missreid}, and dynamic modality selection during inference for incomplete classification~\cite{du2026inference}. These methods mainly focus on recovering or selecting useful modality information for recognition tasks. In contrast, APRTrack separates whole modality degradation and local target corruption in RGB-Event tracking, and models them through hierarchical perturbation and controlled historical retrieval.

\subsection{Memory-based Retrieval Networks}
Memory mechanisms have been widely used in visual tracking to exploit historical target information beyond the current frame. Existing trackers improve robustness by modeling temporal context, maintaining target candidates, updating dynamic templates, or retrieving historical prompts and tokens~\cite{wang2021transformer,mayer2021learning,cai2024hiptrack,zheng2024odtrack,xie2024autoregressive,kang2025exploring}. In RGB-Event tracking, AMTTrack~\cite{wang2024long} introduces associative memory for long-term frame-event tracking, while MamTrack~\cite{sun2025exploring} and Mamba-FETrack V2~\cite{wang2025mambafetrack} exploit long-range temporal modeling with state space models. Beyond tracking, Modern Hopfield Networks~\cite{ramsauer2020hopfield} connect continuous associative memory with attention, enabling memory retrieval to be viewed as associating current states with stored patterns. Recent studies further revisit Hopfield retrieval from more general or adaptive perspectives, such as Universal Hopfield Networks~\cite{millidge2022universal} and Adaptive Hopfield Networks~\cite{wang2026adaptive}. Most tracking methods use memory mainly for temporal aggregation or template adaptation, while retrieval reliability under locally corrupted observations remains less explored. FCHR instead retrieves historical target cues for local compensation by estimating retrieval reliability in the association process.

\section{Our Proposed Approach}

\subsection{Overview}
\begin{figure*}
\centering
\includegraphics[width=\linewidth]{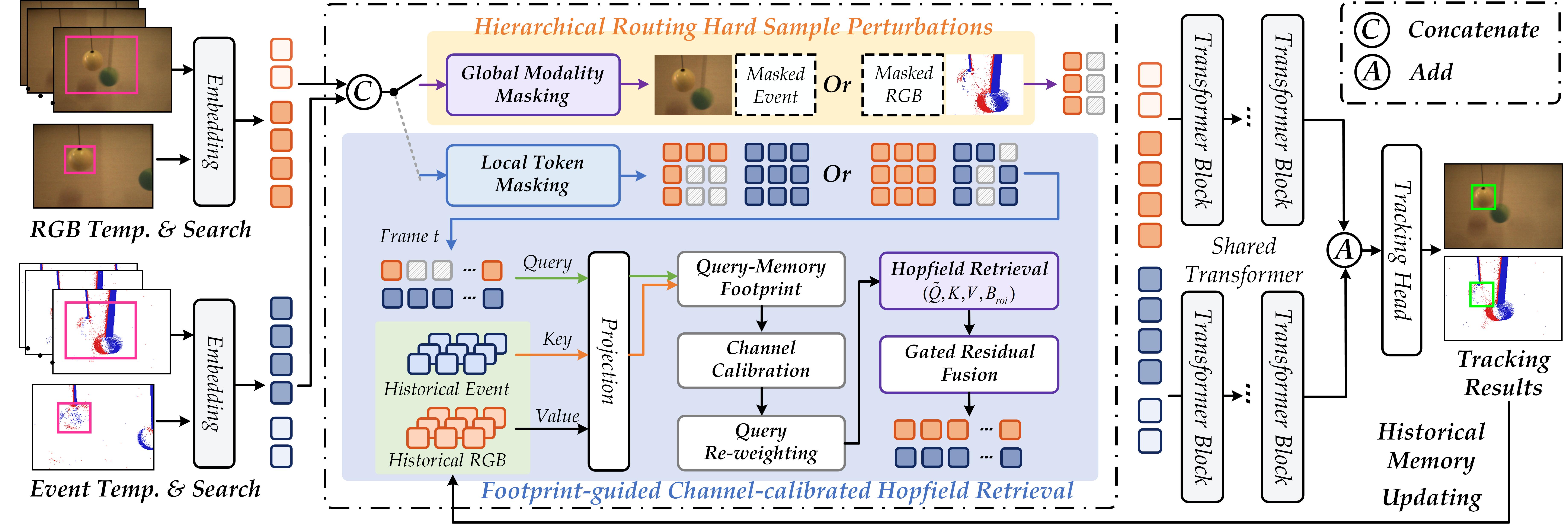}
\caption{An overview of the proposed APRTrack framework for missing-robust RGB-Event tracking. APRTrack first maps RGB and Event template-search inputs into token representations, then applies adversarial hierarchical perturbation and footprint-guided channel-calibrated Hopfield retrieval before the Transformer backbone to model structured degradation and introduce controlled historical compensation. The fused search representation is finally fed into the tracking head for target localization.}
\label{fig:framework}
\end{figure*}

As illustrated in Fig.~\ref{fig:framework}, APRTrack adopts a unified RGB-Event tracking pipeline, which is constructed upon the canonical Transformer-based single-object tracker proposed in prior works \cite{ye2022joint, tang2025revisiting}. Given RGB and Event template-search inputs, the model first maps the four image regions into token representations through shared patch embedding. The adversarial hierarchical perturbation and historical retrieval modules are inserted between patch embedding and the Transformer backbone~\cite{dosovitskiy2020image}, and operate mainly on search tokens to improve robustness to structured missing without changing the backbone tracking pipeline. The processed search tokens are then fed into the shared Transformer backbone together with template tokens for relation modeling. The dual modality fusion module produces the search region features, from which the tracking head predicts classification responses and bounding boxes. At inference time, the perturbation modules are disabled, while historical memory update and FCHR retrieval remain active to use historical target cues for more stable current frame localization. 

\subsection{Preliminary: Modern Hopfield Network} 
Modern Hopfield Networks~\cite{ramsauer2020hopfield}~\footnote{\url{https://github.com/Event-AHU/Awesome_Modern_Hopfield_Networks}} formulate associative memory as a continuous state update process, where an input state converges to a stable state by minimizing an energy function defined over stored patterns. Given stored patterns $\{\bm{y}_i\}_{i=1}^{N}$ and an input state pattern $\bm{r}$, the energy function is written as
\begin{equation}
    E(\bm{r}) = -\frac{1}{\beta}\log\sum_{i=1}^{N}\exp(\beta \bm{y}_i^{\top}\bm{r})
    + \frac{1}{2}\|\bm{r}\|^2 + C,
\end{equation}
where $\beta$ denotes the inverse temperature parameter, $\bm{y}_i$ is the $i$-th stored pattern, $N$ is the number of stored patterns, and $C$ is a constant term independent of $\bm{r}$. This energy form leads to an associative update according to the similarity between the input state and the stored patterns. Let $\bm{Y}=(\bm{y}_1,\dots,\bm{y}_N)^{\top}$ denote the memory bank. The Modern Hopfield update can be written in an attention-like form as
\begin{equation}
    \bm{r}^{(t+1)} = \bm{Y}^{\top}\mathrm{Softmax}\left(\beta \bm{Y}\bm{r}^{(t)}\right),
\end{equation}
where $\bm{r}^{(t)}$ and $\bm{r}^{(t+1)}$ denote the state patterns before and after the update. The softmax weights measure the association strength between the current state and each stored pattern, while $\beta$ controls the sharpness of the softmax distribution.

In deep networks, state patterns and stored patterns are usually projected into a learnable association space. Given query tokens $Q\in\mathbb{R}^{N_q\times D}$, memory keys $K\in\mathbb{R}^{N_m\times D}$, and memory values $V\in\mathbb{R}^{N_m\times D}$, the standard Hopfield association used in this work is defined as
\begin{equation}
    \mathcal{H}(Q,K,V)=\mathrm{Softmax}\left(\beta QK^\top\right)V.
\end{equation}
Here, $Q$ initiates memory access, $K$ determines the association distribution over stored patterns, and $V$ provides the memory content to be aggregated. To introduce target priors during historical retrieval, APRTrack extends this association with a bias term:
\begin{equation}
    \mathcal{H}(Q,K,V;B)=\mathrm{Softmax}\left(\beta QK^\top+B\right)V.
\end{equation}
The current frame tokens are treated as state patterns, while historical RGB and Event tokens serve as stored patterns, allowing the model to retrieve target-related cues from previous frames for cross-temporal compensation. 

Building upon this network, some researchers have improved model performance on pre-trained large models~\cite{furst2022cloob,hu2024outlier,niu2024beyond}, as well as in nuclear fusion~\cite{ma2024exploiting} and time series analysis~\cite{auer2023conformal,wu2024stanhop}. 
Motivated by its associative retrieval capability, we adapt the biased Hopfield association to RGB-Event tracking and introduce Footprint-guided Channel-calibrated Hopfield Retrieval (FCHR) for reliable and controlled historical compensation.

\subsection{Input Representation}
The input of APRTrack consists of template and search images from RGB and Event modalities. Let the RGB template and search images be denoted as $I_z^r \in \mathbb{R}^{C\times H_z\times W_z}$ and $I_x^r \in \mathbb{R}^{C\times H_x\times W_x}$, and the Event template and search images as $I_z^e \in \mathbb{R}^{C\times H_z\times W_z}$ and $I_x^e \in \mathbb{R}^{C\times H_x\times W_x}$, respectively. The Event images are generated by stacking event streams within a fixed temporal interval, where each event is represented as $e_j=(x_j,y_j,t_j,p_j)$ with spatial coordinates ($x_j,y_j$), timestamp $t_j$, and polarity $p_j$. 
After shared patch embedding and positional encoding, the four inputs are mapped into RGB/Event template tokens and search tokens:
\begin{equation}
    \mathcal{T}=\{Z^r, Z^e\}, \quad \mathcal{X}=\{X^r, X^e\}.
\end{equation}
For each modality $m\in\{r,e\}$, the template and search tokens satisfy $Z^m \in \mathbb{R}^{N_z \times D}$ and $X^m \in \mathbb{R}^{N_x \times D}$, respectively, where $D$ denotes the token dimension, $N_z=\frac{H_z W_z}{P^2}$ and $N_x=\frac{H_x W_x}{P^2}$ denote the numbers of template and search tokens, and $P$ denotes the patch size. This representation serves as input to the following modules, where hierarchical perturbation and retrieval operate primarily on the search tokens $\mathcal{X}$.

\subsection{Adversarial Hierarchical Perturbation Module} 
The adversarial hierarchical perturbation module explicitly simulates two categories of structured degradation in RGB-Event tracking: modality-level missing and spatial-level missing. 
The former denotes the complete loss of valid observations from either RGB or Event in the current frame, while the latter denotes occlusion, truncation, or contamination that affects only local target regions. Since these two degradation types impose different robustness requirements, APRTrack does not rely on a unified random noise perturbation. Instead, it separately constructs modality-wise and spatial-wise perturbations to simulate full modality dropout and local target occlusion scenarios. 
Its adversarial property stems from the iterative optimization interplay between the perturbation branch and tracking backbone. Guided by gradient reversal~\cite{ganin2015unsupervised}, the perturbation branch generates harder missing-state samples, whereas the tracking backbone is forced to learn discriminative and robust feature representations against such corruptions. This hierarchical design matches the simulated degradation patterns with real-world missing-modality semantics, while also delineating explicit module partitions to facilitate the hierarchical routing strategy introduced afterwards.

$\bullet$ \textbf{Global Modality-level Adversarial Perturbation.~}
The modality-level perturbation models whole modality absence. Let the patch-embedded RGB and Event search tokens be denoted as $X^r$ and $X^e$, respectively. To simulate the case where one modality fails while the other remains available, we define the mutually exclusive modality perturbation operator $\mathcal{P}_{\mathrm{mod}}$ as
\begin{equation}
    \mathcal{P}_{\mathrm{mod}}(X^r, X^e) = (\alpha^r X^r, \alpha^e X^e).
\end{equation}
Here, $(\alpha^r,\alpha^e) \in \{(1,1),(0,1),(1,0)\}$, where $(1,1)$ denotes dual modality preservation, $(0,1)$ denotes RGB missing, and $(1,0)$ denotes Event missing. This mutual exclusion constraint prevents both inputs from being removed simultaneously, keeping the training difficulty within a learnable range and encouraging the model to reduce its reliance on complete dual modality observations.
Unlike modality dropout with fixed probabilities, this perturbation does not enumerate missing patterns in advance. It adaptively selects the perturbation form through a learnable adversarial mutually exclusive gate conditioned on the current dual modality state. The modality selection vector is predicted by a gating function with gradient reversal:
\begin{equation}
    \bm{\alpha} = \Psi_{\mathrm{mod}}\big(\mathrm{GRL}(\rho(X^r, X^e))\big),
\end{equation}
where $\bm{\alpha}=(\alpha^r,\alpha^e)$, $\rho(\cdot)$ denotes the dual modality context aggregation function, and $\Psi_{\mathrm{mod}}$ denotes the mutually exclusive modality gate. The Gradient Reversal Layer (GRL) drives the gate to generate more challenging modality-missing states for the current tracking representation, forcing the backbone to learn robust features when only one modality is available. In other words, this branch does not simply create incomplete inputs; it encourages the model to maintain target localization capability under incomplete modality conditions. During training, hard Gumbel-Softmax~\cite{jang2017categorical,maddison2017concrete} is used for discrete selection. A lightweight balance regularization is applied during adversarial training to prevent the gate from collapsing to a single modality state or always avoiding perturbation.

$\bullet$ \textbf{Local Spatial-level Adversarial Continuous Perturbation.~}
The spatial-level perturbation models local target absence. For the spatial grid of a search token set $X$, we define the spatial perturbation operator $\mathcal{P}_{\mathrm{spa}}$, which abstracts local target incompleteness as more realistic continuous region occlusion through adversarial spatial selection:
\begin{equation}
    \mathcal{P}_{\mathrm{spa}}(X; s) = (1 - S_s) \odot X,
\end{equation}
where $S_s$ denotes the continuous occlusion mask induced by the candidate window $s$. The key observation is that local target absence usually appears as spatially continuous structural damage related to the target position, rather than irregular scattered token dropout. Therefore, spatial perturbation should corrupt local observations while keeping the occlusion pattern consistent with real scenes, enabling the model to preserve structural awareness and localization stability when the target is partially missing. Window selection depends on both the spatial response of current tokens and target region constraints. Specifically, the window is determined by a spatial scoring function with gradient reversal:
\begin{equation}
    s = \Psi_{\mathrm{spa}}\big(\mathrm{GRL}(\eta(X)), G\big),
\end{equation}
where $\eta(\cdot)$ denotes token-level spatial scoring and $G$ denotes the soft target mask generated from the target bounding box. 
To avoid spatial perturbations degenerating into semantically meaningless large-scale erasure of the target object, we impose a target overlap constraint on the candidate window scores, where candidate windows with excessive target overlap are penalized during adversarial spatial selection and thus receive lower selection scores.
For training, hard Gumbel-Softmax is adopted to sample continuous rectangular windows, and a retention mask is built via straight-through estimation. This design yields explicit hard occlusion masks in the forward pass while preserving differentiable soft selection gradients for backpropagation. Consequently, the generated spatial perturbations align closely with realistic semantic patterns of target occlusion or local cropping, instead of devolving into unstructured random token dropout.

\subsection{Footprint-guided Channel-calibrated Hopfield Retrieval} 

\begin{figure}[!htp]
\centering
\includegraphics[width=0.95\linewidth]{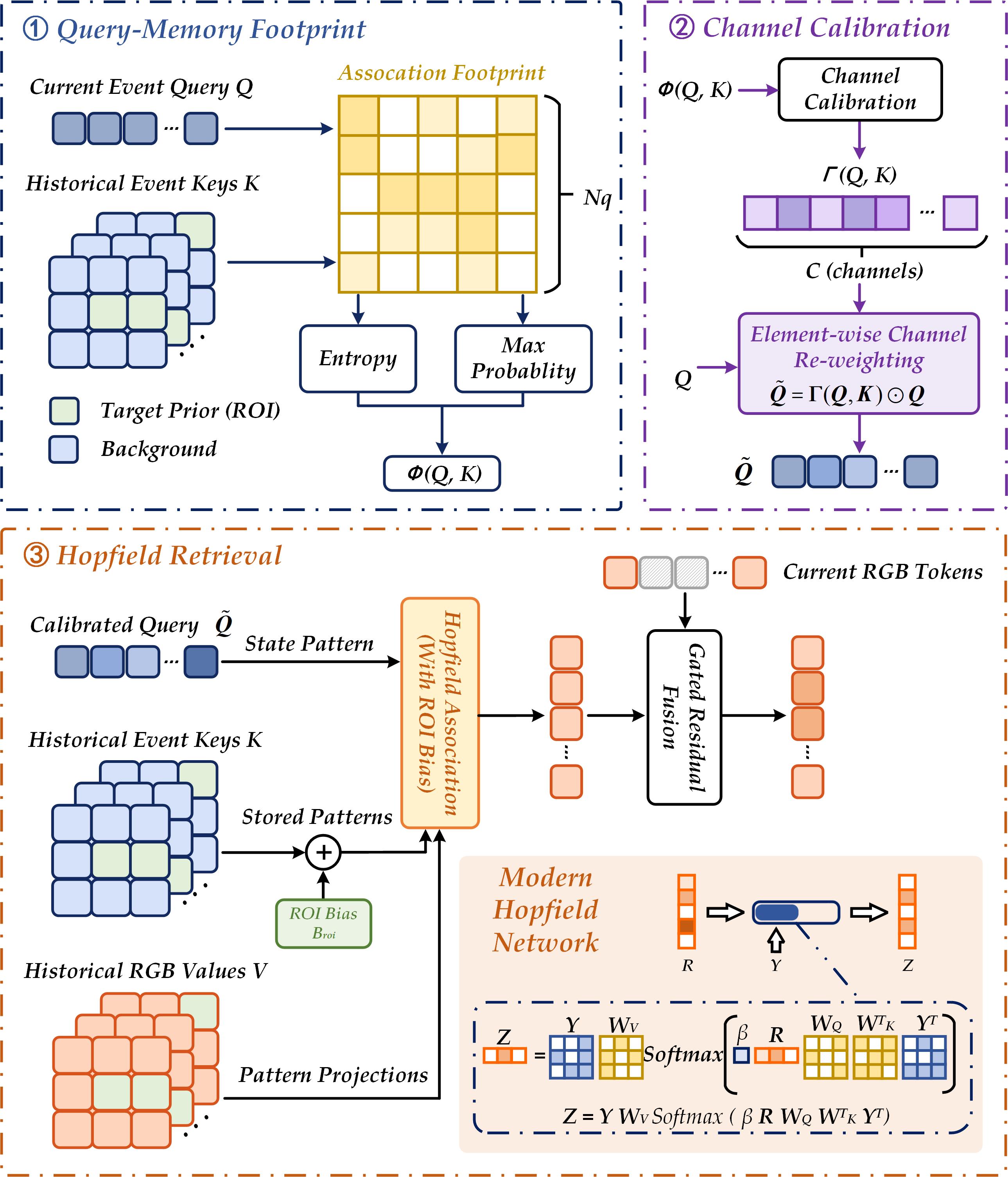}
\caption{Detailed architecture of (1) query-memory association footprint estimation, (2) footprint-guided channel calibration, and (3) ROI-biased Hopfield retrieval with gated residual fusion.}
\label{fig:hopfield}
\end{figure}

In scenarios with local target absence, cross-temporal history often retains cues related to the current target, making historical retrieval potentially useful. 
However, historical information is not inherently reliable. Without target region constraints and retrieval reliability assessment, retrieval may introduce background clutter, outdated appearances, or false matches.
Motivated by this observation, we propose Footprint-guided Channel-calibrated Hopfield Retrieval (FCHR), which aims to provide limited and trustworthy historical compensation under local degradation, rather than unconstrainedly completing the current representation with historical features. 
Instead of using the raw association response as the retrieved feature, FCHR first computes a lightweight query-memory association map, which we refer to as the association footprint because it records how the current query distributes its matches over historical memory. This footprint is then summarized to assess retrieval reliability and calibrate the query channels before Hopfield association.
Toward this goal, FCHR consists of three components: target-constrained historical memory, footprint-guided channel calibration, and gated residual fusion. 

$\bullet$ \textbf{Target-Constrained Historical Memory.~}
FCHR first organizes historical information into target-constrained memory representations, rather than treating historical features as an unstructured feature pool. For historical frames, the module stores RGB tokens, Event tokens, and soft target masks generated from target bounding boxes. The historical memory before frame $t$ is written as
\begin{equation}
    \mathcal{M}_t=\{(K^r_\tau,K^e_\tau,G_\tau)\}_{\tau<t},
\end{equation}
where $K^r_\tau$ and $K^e_\tau$ denote historical RGB and Event tokens, and $G_\tau$ denotes the soft target mask. During retrieval, multi-frame historical tokens are concatenated along the token dimension to form the memory, and a region-of-interest (ROI) bias is constructed from the soft target masks:
\begin{equation}
    B_{\mathrm{roi}} = \gamma (G - \mathbf{1}),
\end{equation}
where biases near target regions approach zero, while background regions receive negative values. The ROI bias is added to the Hopfield association logits, suppressing background memory tokens and encouraging retrieval to focus on target-related historical regions. In this way, the memory organization imposes a target prior on the retrieval scope.

Based on this memory, FCHR employs cross-modal retrieval to generate historical compensation. To complete the RGB representation, the current Event query locates historical target patterns in the Event memory and retrieves the corresponding RGB memory values; conversely, to complete the Event representation, the current RGB query locates historical target patterns in the RGB memory and retrieves the Event memory values. This process is uniformly written as
\begin{equation}
    \begin{gathered}
        \hat{X}^{r}=\mathcal{H}(Q=X^{e},K=K^{e},V=K^{r};B_{\mathrm{roi}}), \\
        \hat{X}^{e}=\mathcal{H}(Q=X^{r},K=K^{r},V=K^{e};B_{\mathrm{roi}}),
    \end{gathered}
\end{equation}
where $\mathcal{H}$ denotes Hopfield association~\cite{ramsauer2020hopfield} with ROI bias. This cross-modal retrieval mechanism leverages the complementary properties of RGB and event streams. The modality with higher instantaneous reliability retrieves target-correlated historical features, while its paired historical counterpart delivers cross-temporal supplementary information to the other branch.

$\bullet$ \textbf{Footprint-Guided Channel Calibration.~}
Target-constrained memory alone is insufficient to guarantee reliable retrieval, as the current query under local occlusion or contamination may have deviated from the true target pattern. To address this, FCHR introduces footprint-guided channel calibration before Hopfield retrieval. It first estimates retrieval reliability from the query-memory association state, and then adjusts the retrieval metric space of the query. Given query $Q$ and memory key $K$, the module computes a lightweight query-memory association footprint:
\begin{equation}
    A_f = \mathrm{Softmax}\left(\frac{(QW_q^f)(KW_k^f)^\top}{\sqrt{d}}\right).
\end{equation}
This footprint is not used as the final retrieval result, but as an intermediate descriptor of retrieval reliability. Specifically, FCHR extracts normalized entropy and maximum matching probability from $A_f$ to form the footprint descriptor:
\begin{equation}
    \phi(Q,K)=\big[\mathcal{E}(A_f); \mathcal{M}(A_f)\big],
\end{equation}
where $\mathcal{E}(\cdot)$ denotes normalized entropy, measuring the dispersion of the association distribution, and $\mathcal{M}(\cdot)$ denotes the maximum association probability, measuring the strongest historical match. The channel calibration function then generates reliability weights from this descriptor and reshapes the query metric space before Hopfield association:
\begin{equation}
    \Gamma(Q,K)=g(\phi(Q,K)), \quad \tilde{Q}=\Gamma(Q,K)\odot Q.
\end{equation}
The calibrated query then enters the Hopfield association with ROI bias:
\begin{equation}
    \hat{X}=\mathcal{H}(\tilde{Q},K,V;B_{\mathrm{roi}})
    = \mathrm{Softmax}\left(\beta \tilde{Q}K^\top + B_{\mathrm{roi}}\right)V.
\end{equation}
Thus, FCHR models historical compensation as a degradation-aware associative memory access process. Historical target memories serve as stored patterns that preserve cross-temporal target representations, the current query affected by local absence acts as the state pattern that initiates retrieval, and cross-modal values generate compensation for the current representation through pattern projection. Footprint-guided channel calibration adjusts the metric space of the state pattern before association, making it more sensitive to channel responses consistent with reliable historical target patterns and reducing associations between contaminated queries and erroneous historical patterns.
In this process, target-constrained memory bias restricts the effective access scope of stored patterns, footprint-guided calibration improves the retrieval reliability of the state pattern, and gated residual fusion further controls the injection strength of pattern projection. Together, these components form the controlled historical compensation mechanism of FCHR.

$\bullet$ \textbf{Gated Residual Fusion.~}
After target-constrained retrieval and footprint-guided channel calibration, FCHR further controls how historical compensation is fused with the current representation. Historical retrieval results are not used to overwrite current tokens directly, but are injected as reliability conditioned compensatory signals. Let the retrieval compensation be denoted as $\hat{X}$ and the current token as $X$. FCHR adopts the following gated residual fusion form:
\begin{equation}
    Y = X + \Omega(X,\hat{X})\odot(\hat{X}-X),
\end{equation}
where $\Omega(\cdot)$ denotes the reliability gating function, which adaptively controls the injection strength of historical information according to the consistency between the current representation and the retrieval compensation.
In practice, the gating function estimates a token-wise gate from the current token, the retrieval residual, and their interaction terms. When historical retrieval is consistent with the current target state, the gate amplifies effective compensation; when the retrieval result is unreliable, the gate suppresses historical information at each token and prevents erroneous memories from overwriting current observations.

\subsection{Hierarchical Routing Training Strategy}
Although modality-level and spatial-level perturbations correspond to whole modality failure and local target absence, respectively, directly applying both to the same sample would excessively weaken the effective information in the current frame. In particular, when one modality has already failed while the target also suffers local occlusion, the model may face an irrecoverable information gap, making robust learning unstable. To address this issue, APRTrack adopts a hierarchical routing training strategy that decouples different degradation pressures at the sample level, rather than applying multiple degradations to the same sample.

During training, each search sample is assigned to one of three branches: clean, modality, and spatial. Let the routing variable be $b_i\in\{\mathrm{clean},\mathrm{mod},\mathrm{spa}\}$. Given $b_i$, the processing form of the search token set $\mathcal{X}_i$ is defined as
\begin{equation}
    \mathcal{P}(\mathcal{X}_i;b_i)=
    \begin{cases}
        \mathcal{X}_i, & b_i=\mathrm{clean}, \\
        \mathcal{P}_{\mathrm{mod}}(\mathcal{X}_i), & b_i=\mathrm{mod}, \\
        \mathcal{P}_{\mathrm{spa}}(\mathcal{X}_i), & b_i=\mathrm{spa}.
    \end{cases}
\end{equation}
The clean branch keeps the original search representation unchanged, providing an unperturbed anchor for normal tracking. The modality and spatial branches expose the model to whole modality missing and local target missing, respectively. This routing design preserves clean representation learning while avoiding the compounded destruction caused by directly superimposing the two degradation types.

Beyond the perturbation itself, historical compensation is also constrained by the routing state. This design assigns clear responsibilities to the three branches: the clean branch maintains a stable current frame representation anchor, the modality branch preserves the single modality robustness objective, and the spatial branch learns to use historical cues for compensation under local target missing. By constraining the scope of historical retrieval through routing, the model avoids using history to buffer whole modality missing pressure while learning controlled compensation under local missing conditions.

Furthermore, to alleviate the impact of early-stage modality-level missing on backbone optimization during joint training, APRTrack adopts a progressive modality perturbation intensity schedule in the modality branch. This schedule is used only to stabilize joint training: optimization starts with milder modality-missing pressure and gradually restores the full perturbation intensity, allowing modality-level perturbation, spatial-level perturbation, and historical retrieval to coexist stably.

\subsection{Tracking Head and Loss Function}
The fused search tokens from the Transformer backbone are fed into the tracking head for target localization. The tracking head follows the standard prediction paradigm of OSTrack~\cite{ye2022joint}, producing target classification responses and bounding box coordinates. During training, the overall objective for each sample consists of bounding box regression and classification losses:
\begin{equation}
    \mathcal{L}=\lambda_{L_1}\mathcal{L}_{1}+\lambda_{GIoU}\mathcal{L}_{GIoU}+\lambda_{focal}\mathcal{L}_{focal},
\end{equation}
where $\mathcal{L}_{1}$ constrains box regression, $\mathcal{L}_{GIoU}$~\cite{rezatofighi2019generalized} measures the geometric overlap between predicted and ground-truth boxes, and $\mathcal{L}_{focal}$~\cite{law2018cornernet} optimizes target response classification.
We set $\lambda_{L_1}=5.0$, $\lambda_{GIoU}=2.0$, and $\lambda_{focal}=1.0$. 

\section{Experiments}

\subsection{Datasets and Evaluation Metric}
To validate the effectiveness of the proposed method, we conduct experiments on four RGB-Event single object tracking benchmarks: \textbf{FE108}~\cite{zhang2021object}, \textbf{COESOT}~\cite{tang2025revisiting}, \textbf{VisEvent}~\cite{wang2023visevent}, and \textbf{FELT}~\cite{wang2024long}. These datasets cover short and long tracking scenarios, and include challenging factors such as illumination variation, fast motion, occlusion, and out-of-view motion. They therefore provide a reliable evaluation of RGB-Event tracking performance under different degradation conditions.

For evaluation, we adopt three standard tracking metrics: \textbf{Success Rate (SR)}, \textbf{Precision Rate (PR)}, and \textbf{Normalized Precision Rate (NPR)}. Specifically, SR measures the overlap quality between predicted and ground-truth bounding boxes, PR reflects target center localization accuracy, and NPR further normalizes center point errors across different target scales.

\subsection{Implementation Details}
For fair comparison, APRTrack is trained separately using the training split of each benchmark dataset and evaluated on its corresponding test split. In the training phase, the basic optimization settings follow OSTrack~\cite{ye2022joint}, and the model is trained with the AdamW optimizer~\cite{loshchilov2017decoupled}. The learning rate is set to 1e-4, the weight decay is set to 1e-4, and the batch size is set to 16. The model is trained for 50 epochs, and the learning rate is decayed by a factor of 0.1 at epoch 40. We adopt HiViT-B~\cite{zhang2023hivit} as the Transformer encoder, with template and search input sizes of $128\times128$ and $256\times256$, respectively. The backbone is initialized from a pretrained MAE-HiViT-B model, while the newly introduced modules are randomly initialized. The hierarchical perturbation and FCHR modules are inserted between the patch embedding and the Transformer backbone. Hierarchical perturbation and routing are enabled during training; at inference time, perturbation modules are disabled, while FCHR and historical memory updates remain active for current frame compensation.
All experiments are conducted on a computing server equipped with an NVIDIA GeForce RTX 4090 GPU. FPS is consistently evaluated on an NVIDIA GeForce RTX 2080 Ti GPU. More details can be found in our source code.

\subsection{Comparison on Public Benchmark Datasets}

$\bullet$ \textbf{Results on FE108 Dataset.~}
\begin{table*}[tbp]
\centering
\footnotesize
\setlength{\tabcolsep}{0.6mm}
\caption{Experimental results (SR/PR) on FE108 dataset.}
\label{tab:fe108_results}
\begin{tabular}{cccccccc}
\toprule
\textbf{SiamBAN~\cite{chen2020siamese}} & \textbf{SiamFC++~\cite{xu2020siamfc++}} & \textbf{KYS~\cite{bhat2020know}} & \textbf{CLNet~\cite{dong2020clnet}} & \textbf{CMT-MDNet~\cite{wang2023visevent}} & \textbf{ATOM~\cite{danelljan2019atom}} & \textbf{DiMP~\cite{bhat2019learning}}\\
22.5/37.4 & 23.8/39.1 & 26.6/41.0 & 34.4/55.5 & 35.1/57.8 & 46.5/71.3 & 52.6/79.1\\
\hline
\textbf{PrDiMP~\cite{danelljan2020probabilistic}} & \textbf{CMT-ATOM~\cite{wang2023visevent}} & \textbf{CEUTrack~\cite{tang2025revisiting}} & \textbf{ViPT~\cite{zhu2023visual}} & \textbf{MamTrack~\cite{sun2025exploring}} & \textbf{AMTTrack~\cite{wang2024long}} & \textbf{Ours} \\
53.0/80.5 & 54.3/79.4 & 55.6/84.5 & 65.8/93.8 & \textbf{66.4}/94.2 & 65.6/95.9 & 65.1/\textbf{97.0} \\
\bottomrule
\end{tabular}
\end{table*}
FE108~\cite{zhang2021object} contains 108 videos with 208,672 frames in total, averaging about 1,932 frames per sequence, and covers 21 target categories with four challenging factors. As shown in Table~\ref{tab:fe108_results}, APRTrack achieves the best PR score of 97.0 among all compared methods and obtains a competitive SR of 65.1. Compared with AMTTrack~\cite{wang2024long}, APRTrack improves PR from 95.9 to 97.0 while maintaining comparable SR. Although its SR is slightly lower than the best result of 66.4 achieved by MamTrack~\cite{sun2025exploring}, APRTrack provides the most accurate target center localization. These results show that APRTrack maintains strong robustness in long tracking scenarios on FE108 while achieving the highest localization precision among the compared trackers.

$\bullet$ \textbf{Results on COESOT Dataset.~}
\begin{table}
\centering
\small
\caption{Experimental Results on COESOT Dataset.}
\label{tab:coesot_results}
\begin{tabular}{l|c|cc}
\hline \toprule [0.5 pt]
\textbf{Trackers} & \textbf{Source} &\textbf{SR}  & \textbf{PR} \\
\hline
\textbf{01. Stark~\cite{yan2021learning}} & ICCV21 & 56.0 & 67.7 \\
\textbf{02. KeepTrack\cite{mayer2021learning}} & ICCV21 & 59.6 & 70.9 \\
\textbf{03. TrDiMP~\cite{wang2021transformer}} & CVPR21 & 60.1 & 72.2 \\
\textbf{04. TransT~\cite{chen2021transformer}} & ECCV22 & 60.5 & 72.4 \\
\textbf{05. OSTrack~\cite{ye2022joint}} & ECCV22 & 59.0 & 70.7 \\
\textbf{06. AiATrack~\cite{gao2022aiatrack}} & ECCV22 & 59.0 & 72.4 \\
\textbf{07. MixFormer~\cite{cui2022mixformer}} & CVPR22 & 55.7 & 66.3 \\
\textbf{08. ToMP101\cite{mayer2022transforming}} & CVPR22 & 59.9 & 71.6 \\
\textbf{09. MDNet~\cite{wang2023visevent}} & TCYB23 & 53.3 & 66.5 \\
\textbf{10. ViPT~\cite{zhu2023visual}} & CVPR23 & 68.3 & 81.0 \\
\textbf{11. SDSTrack~\cite{hou2024sdstrack}} & CVPR24 & 66.7 & 79.7 \\
\textbf{12. UnTrack~\cite{wu2024single}} & CVPR24 & 67.9 & 80.9 \\
\textbf{13. CEUTrack\cite{tang2025revisiting}} & PR25 & 62.7 & 76.0 \\
\textbf{14. CMDTrack~\cite{zhang2025cross}} & TPAMI25 & 65.7 & 74.8 \\
\textbf{15. LMTrack~\cite{xu2025less}} & AAAI25 & 58.4 & 71.1 \\
\textbf{16. MCITrack~\cite{kang2025exploring}} & AAAI25 & 64.7 & 78.1 \\
\textbf{17. AMTTrack~\cite{wang2024long}} & arXiv25 & \textbf{68.8} & 82.9 \\
\textbf{18. SpikeFET~\cite{yang2026fully}} & NeurIPS25 & 68.5 & 81.7 \\
\textbf{19. UTPTrack-O~\cite{wu2026utptrack}} & CVPR26 & 57.8 & 70.2 \\
\textbf{20. UTPTrack-S~\cite{wu2026utptrack}} & CVPR26 & 64.7 & 77.6 \\
\textbf{21. LASTracker~\cite{wang2026lastracker}} & PR26 & 62.6 & 72.3 \\
\hline
\textbf{22. Ours} & - & 68.3 & \textbf{84.0} \\
\hline \toprule [0.5 pt]
\end{tabular}
\end{table}
COESOT~\cite{tang2025revisiting} is a large-scale RGB-Event tracking benchmark with 1,354 video sequences, 478,721 frames, and 90 target categories, covering diverse scenes such as indoor environments, streets, and zoos. As shown in Table~\ref{tab:coesot_results}, APRTrack achieves the best PR score of 84.0, outperforming AMTTrack~\cite{wang2024long} with 82.9 and ViPT~\cite{zhu2023visual} with 81.0. It also obtains an SR score of 68.3, close to the best result of 68.8 achieved by AMTTrack and higher than most existing trackers. This indicates that APRTrack provides more accurate target localization on COESOT without sacrificing overlap performance. These results show the effectiveness of the proposed missing-robust representation learning framework in large-scale RGB-Event tracking scenarios.

$\bullet$ \textbf{Results on VisEvent Dataset.~}
\begin{table}
\centering
\small
\caption{Experimental Results on VisEvent Dataset.}
\label{tab:visevent_results}
\begin{tabular}{l|c|cc}
\hline \toprule [0.5 pt]
\textbf{Trackers} & \textbf{Source} &\textbf{SR}  & \textbf{PR} \\
\hline
\textbf{01. ATOM~\cite{danelljan2019atom}} & CVPR19 & 41.2 & 60.8 \\
\textbf{02. DiMP50~\cite{bhat2019learning}} & ICCV19 & 45.1 & 66.1 \\
\textbf{03. SiamCAR~\cite{guo2020siamcar}} & CVPR20 & 42.0 & 59.9 \\
\textbf{04. PrDiMP50~\cite{danelljan2020probabilistic}} & CVPR20 & 45.3 & 64.4 \\
\textbf{05. SiamR-CNN~\cite{voigtlaender2020siam}} & CVPR20 & 49.9 & 65.9 \\
\textbf{06. Stark~\cite{yan2021learning}} & ICCV21 & 44.6 & 61.2 \\
\textbf{07. TransT~\cite{chen2021transformer}} & ECCV22 & 47.4 & 65.0 \\
\textbf{08. OSTrack~\cite{ye2022joint}} & ECCV22 & 53.4 & 69.5 \\
\textbf{09. MDNet~\cite{wang2023visevent}} & TCYB23 & 42.6 & 66.1 \\
\textbf{10. ViPT~\cite{zhu2023visual}} & CVPR23 & 59.2 & 75.8 \\
\textbf{11. SDSTrack~\cite{hou2024sdstrack}} & CVPR24 & 59.7 & 76.7 \\
\textbf{12. UnTrack~\cite{wu2024single}} & CVPR24 & 59.7 & 76.3 \\
\textbf{13. AMTTrack~\cite{wang2024long}} & arXiv25 & 60.1 & 78.1 \\
\textbf{14. SpikeFET~\cite{yang2026fully}} & NeurIPS25 & 59.0 & 75.3 \\
\textbf{15. SEATrack~\cite{su2026seatrack}} & CVPR26 & \textbf{60.3} & 77.1 \\
\textbf{16. LASTracker~\cite{wang2026lastracker}} & PR26 & 52.9 & 70.4 \\
\hline
\textbf{17. Ours} & - & 60.0 & \textbf{79.4} \\
\hline \toprule [0.5 pt]
\end{tabular}
\end{table}
VisEvent~\cite{wang2023visevent} contains 820 video sequences and 371,127 frames, with an average length of 453 frames per sequence, and covers 17 challenging factors across RGB and Event modalities. As shown in Table~\ref{tab:visevent_results}, APRTrack achieves the best PR score of 79.4, outperforming SDSTrack~\cite{hou2024sdstrack} with 76.7, AMTTrack~\cite{wang2024long} with 78.1, and UnTrack~\cite{wu2024single} with 76.3. Although its SR score of 60.0 is slightly lower than the 60.3 achieved by SEATrack~\cite{su2026seatrack}, it remains close to the leading method. This indicates that APRTrack provides more accurate target center localization while preserving competitive success performance on VisEvent. These results further show that the proposed framework generalizes well to complex short-term RGB-Event tracking scenarios.

$\bullet$ \textbf{Results on FELT Dataset.~}
\begin{table}
\centering
\small
\caption{Experimental results on FELT dataset.}
\label{tab:felt_results}
\begin{tabular}{l|c|ccc}
\toprule
\textbf{Trackers} & \textbf{Source} &\textbf{SR}  &\textbf{PR} &\textbf{NPR}\\
\hline
\textbf{01. Stark~\cite{yan2021learning}} & ECCV22 & 52.7 & 67.9 & 62.8 \\
\textbf{02. OSTrack~\cite{ye2022joint}} & ECCV22 & 52.3 & 65.9 & 63.3 \\
\textbf{03. MixFormer~\cite{cui2022mixformer}} & CVPR22 & 53.0 & 67.5 & 63.8 \\
\textbf{04. AiATrack~\cite{gao2022aiatrack}} & ECCV22 & 52.2 & 66.7 & 62.8 \\
\textbf{05. SimTrack~\cite{chen2022backbone}} & ECCV22 & 49.7 & 63.6 & 59.82 \\
\textbf{06. GRM~\cite{gao2023generalized}} & CVPR23  & 52.1 & 65.6 & 62.9 \\
\textbf{07. ROMTrack~\cite{cai2023robust}} & ICCV23  & 51.8 & 65.8 & 62.7 \\
\textbf{08. ViPT~\cite{zhu2023visual}} & CVPR23 & 52.8 & 65.3 & 63.1  \\
\textbf{09. SeqTrack~\cite{chen2023seqtrack}} & CVPR23  & 52.7 & 66.9 & 63.4 \\
\textbf{10. ARTrackv2~\cite{bai2024artrackv2}} & CVPR24 & 52.3 & 65.2 & 62.8 \\
\textbf{11. HIPTrack~\cite{cai2024hiptrack}} & CVPR24 & 51.6 & 65.6 & 62.2 \\
\textbf{12. ODTrack~\cite{zheng2024odtrack}} & AAAI24 & 52.2 & 66.0 & 63.5 \\
\textbf{13. EVPTrack~\cite{shi2024explicit}} & AAAI24 & 53.8 & 68.7 & 64.8 \\
\textbf{14. AQATrack~\cite{xie2024autoregressive}} & CVPR24 & 54.0 & 69.1 & 64.7 \\
\textbf{15. SDSTrack~\cite{hou2024sdstrack}} & CVPR24 & 53.7 & 66.4 & 64.1 \\
\textbf{16. UnTrack~\cite{wu2024single}} & CVPR24 & 53.6 & 66.0 & 63.9 \\
\textbf{17. FERMT~\cite{zheng2024exploring}} & ECCV24 & 51.8 & 66.1 & 62.9 \\
\textbf{18. LMTrack~\cite{xu2025less}} & AAAI25 & 50.9 & 63.9 & 61.8 \\
\textbf{19. AsymTrack~\cite{zhu2025two}} & AAAI25 & 51.9 & 66.7 & 62.0 \\
\textbf{20. ORTrack~\cite{wu2025learning}} & CVPR25 & 48.4 & 61.7 & 59.2 \\
\textbf{21. UNTrack~\cite{qin2025must}} & CVPR25 & 50.0 & 63.9 & 61.6 \\
\textbf{22. SUTrack~\cite{Chen2024SuTrack}} & AAAI25 & 55.1 & 68.9 & 65.2 \\
\textbf{23. XTrack~\cite{Tan2025bXTrack}} & ICCV25 & 55.1 & 67.6 & 65.3 \\
\textbf{24. AMTTrack~\cite{wang2024long}} & arXiv25 & 54.8 & 67.9 & 65.7 \\
\textbf{25. UTPTrack-O~\cite{wu2026utptrack}} & CVPR26 & 51.9 & 65.1 & 62.9 \\
\textbf{26. UTPTrack-S~\cite{wu2026utptrack}} & CVPR26 & 54.7 & 68.5 & 65.2 \\
\textbf{27. SpikeTrack~\cite{zhang2026spiketrack}} & CVPR26 & 52.5 & 67.9 & 63.5 \\
\hline
\textbf{28. Ours} & - & \textbf{55.3} & \textbf{70.1} & \textbf{66.6} \\
\bottomrule
\end{tabular}
\end{table}
FELT~\cite{wang2024long} is a large-scale long-term Frame-Event tracking dataset with 1,044 videos, 1,949,680 frames, and 60 target categories, covering indoor and outdoor scenarios with 14 annotated challenging attributes, such as fast motion, low illumination, overexposure, and out-of-view motion. As shown in Table~\ref{tab:felt_results}, APRTrack achieves the best performance on FELT, with 55.3 SR, 70.1 PR, and 66.6 NPR. Compared with AMTTrack~\cite{wang2024long}, APRTrack improves SR, PR, and NPR by 0.5, 2.2, and 0.9 points, respectively. It also outperforms recent strong trackers such as SUTrack~\cite{Chen2024SuTrack} and XTrack~\cite{Tan2025bXTrack}, which obtain similar SR scores but lower PR and NPR. These results show that APRTrack improves RGB-Event tracking accuracy and provides more stable target localization in long-term scenarios with complex missing, occlusion, and appearance variation.

\subsection{Ablation Study}
\noindent $\bullet$ \textbf{Component Analysis.~}
\begin{table}[tb]
\centering
\small
\setlength{\tabcolsep}{1.3mm}
\caption{Component analysis on COESOT Dataset.}
\label{tab:ablation_components}
\begin{tabular}{c|cccc|ccc|cc}
\toprule
\textbf{\#} & \textbf{MPL} & \textbf{SPL} & \textbf{HR} & \textbf{FC} & \textbf{SR} & \textbf{PR} & \textbf{NPR} & \textbf{Params} & \textbf{FLOPs} \\
\hline
1 & \null & \null & \null & \null & 66.3 & 81.2 & 79.4 & 70.65M & 56.87G \\
2 & \ding{51} & \null & \null & \null & 67.8 & 83.3 & 81.4 & 70.79M & 56.87G  \\
3 & \null & \ding{51} & \null & \null & 66.2 & 82.0 & 79.9 & 70.66M & 56.87G \\
4 & \null & \ding{51} & \ding{51} & \null & 66.5 & 82.5 & 80.7 & 80.65M & 78.22G \\
5& \null & \ding{51} & \ding{51} & \ding{51} & 67.5 & 83.6 & 81.7 & 81.73M & 78.50G \\
6 &\ding{51} & \ding{51} & \ding{51} & \ding{51} & \textbf{68.3} & \textbf{84.0} & \textbf{82.0} & 81.86M & 78.50G \\
\bottomrule
\end{tabular}
\end{table}
As shown in Table~\ref{tab:ablation_components}, we ablate four key components of APRTrack: modality-level perturbation layer (MPL), spatial-level perturbation layer (SPL), Hopfield retrieval (HR), and footprint-channel calibration (FC). Each component contributes to the tracking performance from a different degradation perspective. Compared with the baseline, adding MPL alone brings gains of 1.5 SR, 2.1 PR, and 2.0 NPR, showing that modality-level adversarial perturbation strengthens the robustness of the RGB-Event tracker against whole modality absence. SPL alone mainly improves localization-related metrics, with gains of 0.8 PR and 0.5 NPR, indicating that spatial-level perturbation introduces useful training difficulty for local target missing, although the overall improvement remains limited without historical compensation. Adding vanilla Hopfield retrieval on top of SPL further improves SR, PR, and NPR by 0.3, 0.5, and 0.8, respectively, suggesting that historical association provides complementary cues for locally corrupted observations.

With footprint-channel calibration, the model gains another 1.0 SR, 1.1 PR, and 1.0 NPR over vanilla Hopfield retrieval. This verifies that the query-memory association footprint helps characterize retrieval reliability and calibrate the channel-wise retrieval metric space for corrupted queries before Hopfield association. Finally, the complete model improves the baseline by 2.0 SR, 2.8 PR, and 2.6 NPR and achieves the best overall performance. These results show that the four components are complementary: MPL targets whole modality missing, SPL models local spatial absence, HR introduces historical association, and FC improves retrieval reliability. Under the hierarchical routing training framework, they jointly improve missing-robust RGB-Event tracking.

\noindent $\bullet$ \textbf{Analysis on MPL Design.~}
\begin{table}[tb]
\centering
\small
\setlength{\tabcolsep}{1.5mm}
\caption{Analysis of MPL design on COESOT Dataset.}
\label{tab:ablation_mpl_design}
\begin{tabular}{l|c|ccc}
\toprule
\textbf{Setting} & \textbf{Selection} & \textbf{SR}  & \textbf{PR} & \textbf{NPR} \\
\hline
Random dropout & random & 67.5 & 82.8 & 81.0 \\
MPL w/o reg. & adv. gate & 67.7 & 82.9 & 81.2 \\
MPL & adv. gate + bal. reg. & \textbf{67.8} & \textbf{83.3} & \textbf{81.4} \\
\bottomrule
\end{tabular}
\end{table}
This experiment analyzes how MPL selects modality missing states. Random dropout serves as a non-adaptive baseline that drops one modality with a fixed probability, while MPL w/o reg. uses an adversarial gate to select more challenging missing states according to the current RGB-Event observation. Without constraints, however, the gate may favor a small subset of modality states, especially since Event observations are usually sparser than RGB in many scenes, leading to imbalanced training difficulty. The complete MPL further introduces balance regularization (bal. reg. in Table~\ref{tab:ablation_mpl_design}) to constrain the selection distribution of different modality missing states and mitigate gate collapse. With balance regularization, MPL improves SR, PR, and NPR over MPL w/o reg. by 0.1, 0.4, and 0.2 points, respectively, and achieves the best results on all three metrics. This shows that adversarial selection and balance regularization jointly improve modality-level missing robustness learning.

\begin{table}[tb]
\centering
\small
\caption{Analysis of spatial perturbation and retrieval designs on COESOT Dataset.}
\label{tab:ablation_designs}
\begin{tabular}{l|ccc}
\toprule
\rowcolor{mygray}\textbf{\# Occlusion Severity} & \textbf{SR} & \textbf{PR} & \textbf{NPR} \\
1.~[0.10, 0.20] & 67.1 & 82.7 & 81.1 \\
\textbf{2.}~\textbf{[0.15, 0.30]} & \textbf{67.5} & \textbf{83.6} & \textbf{81.7} \\
3.~[0.20, 0.40] & 66.4 & 82.5 & 80.6 \\
\hline
\rowcolor{mygray}\textbf{\# Search Number} & \textbf{SR} & \textbf{PR} & \textbf{NPR} \\
\hline
1.~2 & 66.8 & 82.6 & 80.9 \\
\textbf{2.}~\textbf{3} & \textbf{67.5} & \textbf{83.6} & \textbf{81.7} \\
3.~4 & 66.4 & 82.2 & 80.5 \\
4.~5 & 66.2 & 82.2 & 80.4 \\
\hline
\rowcolor{mygray}\textbf{\# Footprint Descriptor} & \textbf{SR} & \textbf{PR} & \textbf{NPR} \\
\hline
1.~w/o footprint & 66.5 & 82.5 & 80.7 \\
2.~Entropy only & 67.0 & 82.9 & 81.2 \\
3.~Max probability only & 66.9 & 82.5 & 80.8 \\
\textbf{4.}~\textbf{Entropy + Max probability} & \textbf{67.5} & \textbf{83.6} & \textbf{81.7} \\
\bottomrule
\end{tabular}
\end{table}

\noindent $\bullet$ \textbf{Analysis on Spatial Occlusion Severity.~}
Spatial-level continuous perturbation simulates local target absence by applying target-constrained rectangular occlusion, where the occlusion area range controls the strength of local degradation. This range denotes the area ratio of the candidate occlusion window to the full search token grid; for example, [0.15, 0.30] means that the selected window covers 15\% to 30\% of the search tokens. We compare three occlusion severity settings, i.e., [0.10, 0.20], [0.15, 0.30], and [0.20, 0.40]. As shown in Table~\ref{tab:ablation_designs}, the moderate range [0.15, 0.30] achieves the best performance. This setting provides sufficient local-missing pressure while preserving recoverable target cues. In contrast, a smaller range cannot adequately simulate local target absence, whereas a larger range may corrupt excessive target regions and turn local absence into severe target damage, weakening the effective correspondence between historical retrieval and current representation.

\noindent $\bullet$ \textbf{Analysis on Training Historical Context.~}
FCHR relies on historical search frames available during training to learn cross-temporal compensation. We analyze the influence of historical context length by varying the search number in the training data, where $\text{search number}=2$ denotes the current frame with one historical search frame, $\text{search number}=3$ denotes the current frame with two historical search frames, and so on. As shown in Table~\ref{tab:ablation_designs}, $\text{search number}=3$ achieves the best performance, indicating that a moderate historical context is more suitable for stable retrieval learning. Too few historical frames provide insufficient cross-temporal target cues for memory-based compensation, whereas further increasing the search number introduces more appearance variation, localization noise, and background interference, weakening the effective association between the current query and historical memory. Therefore, FCHR does not simply benefit from longer histories; instead, it requires a moderate and reliable temporal context for stable retrieval learning.

\noindent $\bullet$ \textbf{Analysis on Footprint Descriptor Design.~}
FCHR estimates retrieval reliability using the query-memory association footprint and calibrates the channel-wise metric space of the query before Hopfield association. We compare four footprint descriptor designs: without footprint, entropy only, max probability only, and entropy combined with max probability. As shown in Table~\ref{tab:ablation_designs}, combining entropy and max probability achieves the best performance, indicating that the two statistics characterize retrieval states from complementary perspectives. Entropy reflects the dispersion of the association distribution and measures retrieval uncertainty, while max probability captures the strongest historical response and indicates whether a clear target-related memory exists. Either statistic alone describes only one aspect of the retrieval state; their combination jointly models distributional uncertainty and strongest-match confidence, providing a more complete reliability cue for channel calibration and leading to more stable and reliable historical compensation.

\noindent $\bullet$ \textbf{Analysis on Hierarchical Routing Strategy.~}
\begin{table}[tb]
\centering
\small
\setlength{\tabcolsep}{2mm}
\caption{Analysis of hierarchical routing strategy on COESOT Dataset.}
\label{tab:ablation_routing}
\begin{tabular}{l|c|ccc}
\toprule
\textbf{Training Strategy} & \textbf{Variant} & \textbf{SR} & \textbf{PR} & \textbf{NPR} \\
\hline
Stacked perturbation & MPL + SPL & 64.0 & 80.5 & 79.4 \\
\hline
\multirow{3}{*}{Hierarchical routing} & w/o clean branch & 67.6 & 83.4 & 81.5 \\
& w/ clean, w/o prog. & 68.2 & 83.6 & 81.7 \\
& w/ clean and prog. & \textbf{68.3} & \textbf{84.0} & \textbf{82.0} \\
\bottomrule
\end{tabular}
\end{table}
Table~\ref{tab:ablation_routing} validates the necessity of hierarchical routing, where prog. denotes progressive modality scheduling: training starts with milder modality perturbation and gradually restores the full perturbation strength. Directly stacking MPL and SPL exposes the same sample to both whole modality missing and local spatial occlusion, which excessively weakens the effective information in the current frame and leads to clear performance degradation. Hierarchical routing without a clean branch already improves SR, PR, and NPR over the stacked setting by 3.6, 2.9, and 2.1 points, respectively, showing that assigning modality-level missing and spatial-level missing to separate training paths avoids information collapse caused by double degradation. However, when all samples are placed under degraded states, the model still lacks an unperturbed current frame representation anchor. Introducing the clean branch further improves SR, PR, and NPR by 0.6, 0.2, and 0.2 points, respectively, indicating that clean samples help balance normal representation learning with structured degradation training. Finally, the complete strategy outperforms the variant without progressive scheduling, suggesting that gradually increasing modality missing pressure helps stabilize early joint training. These results show that the clean branch and progressive scheduling are important for stably combining MPL, SPL, and historical retrieval compensation.

\noindent $\bullet$ \textbf{Success Rate Under Challenging Attributes.~}
\begin{figure}[tbp]
\centering
\includegraphics[width=0.95\linewidth]{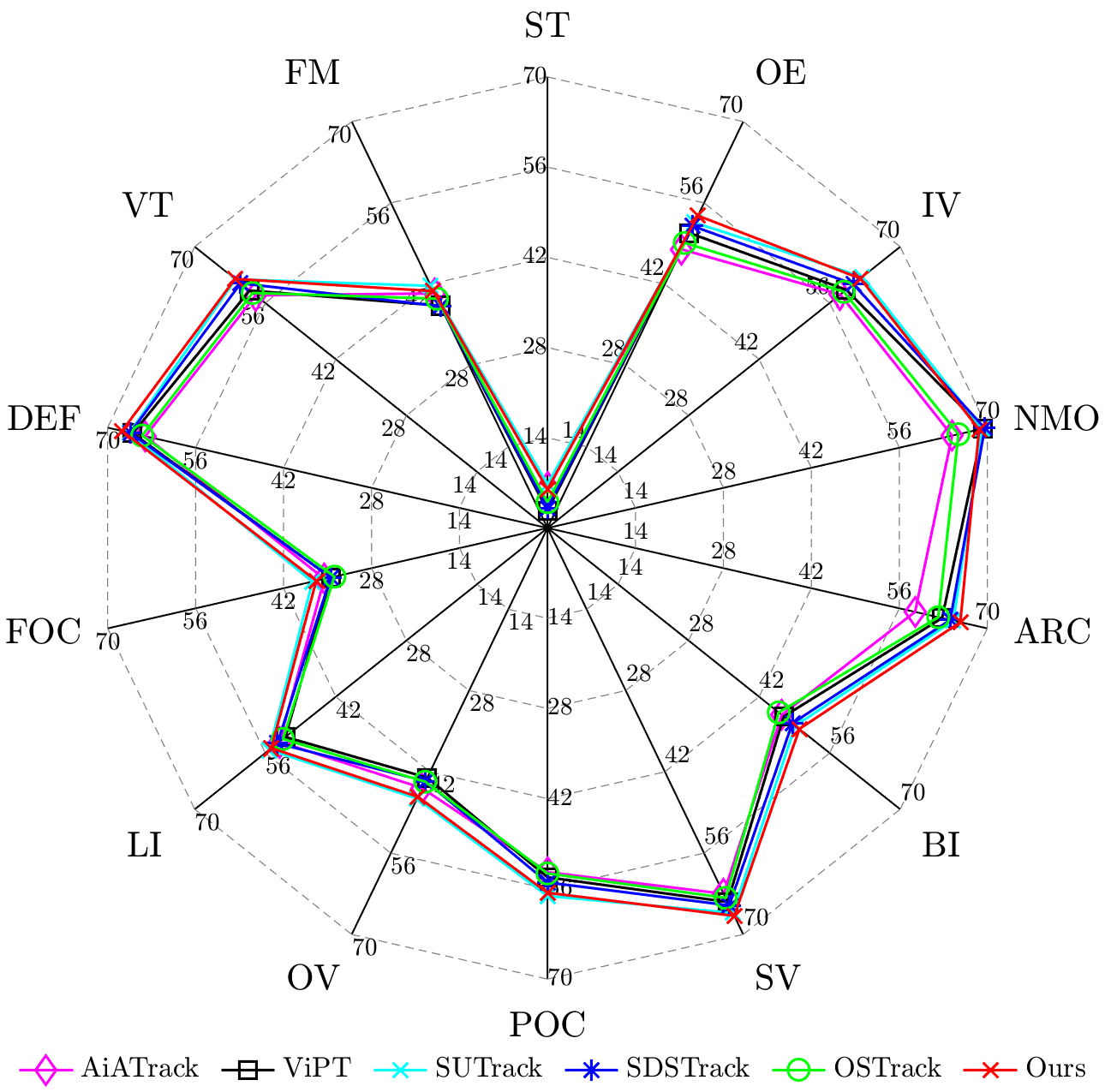}
\caption{Success rate comparison under 14 challenging attributes on FELT.}
\label{fig:sr_attributes}
\end{figure}
To further analyze robustness under different challenging factors, we report the success rate of different trackers on 14 attributes of the FELT dataset, including fast motion (FM), viewpoint transformation (VT), full occlusion (FOC), low illumination (LI), small target (ST), out-of-view (OV), and other challenging factors. As shown in Fig.~\ref{fig:sr_attributes}, APRTrack is compared with OSTrack~\cite{ye2022joint}, SDSTrack~\cite{hou2024sdstrack}, SUTrack~\cite{Chen2024SuTrack}, ViPT~\cite{zhu2023visual}, and AiATrack~\cite{gao2022aiatrack} in terms of attribute-level SR. APRTrack achieves consistent improvements on most attributes, especially under viewpoint transformation, deformation, partial occlusion, scale variation, and aspect ratio change, where the target appearance or geometric state changes substantially. It also remains competitive under stronger degradation factors such as fast motion, full occlusion, low illumination, and out-of-view. The relatively low scores on small target (ST) indicate that this attribute remains highly challenging, since limited target pixels provide insufficient appearance and event cues, making accurate localization difficult for all compared trackers. These results indicate that the gains of APRTrack are not limited to a single attribute, but are reflected across diverse missing, occlusion, and appearance variation conditions.

\subsection{Efficiency Analysis}
As shown in Table~\ref{tab:ablation_components}, APRTrack improves tracking performance with acceptable computational overhead. Since MPL and SPL are training-time perturbation modules, we report the training-time parameter count to reflect the full optimization framework, while FLOPs are measured at inference time. The complete model increases the parameter count from 70.65M to 81.86M, mainly due to the historical compensation and footprint calibration modules. When FCHR is enabled, the inference FLOPs increase from 56.87G to 78.50G, indicating that the additional historical retrieval cost remains moderate. In addition, APRTrack runs at 31 FPS on FELT, showing its practical real-time tracking capability.

\subsection{Visualization}

\noindent $\bullet$ \textbf{Compensation Gate Dynamics.}
\begin{figure}[tbp]
\centering
\includegraphics[width=0.95\linewidth]{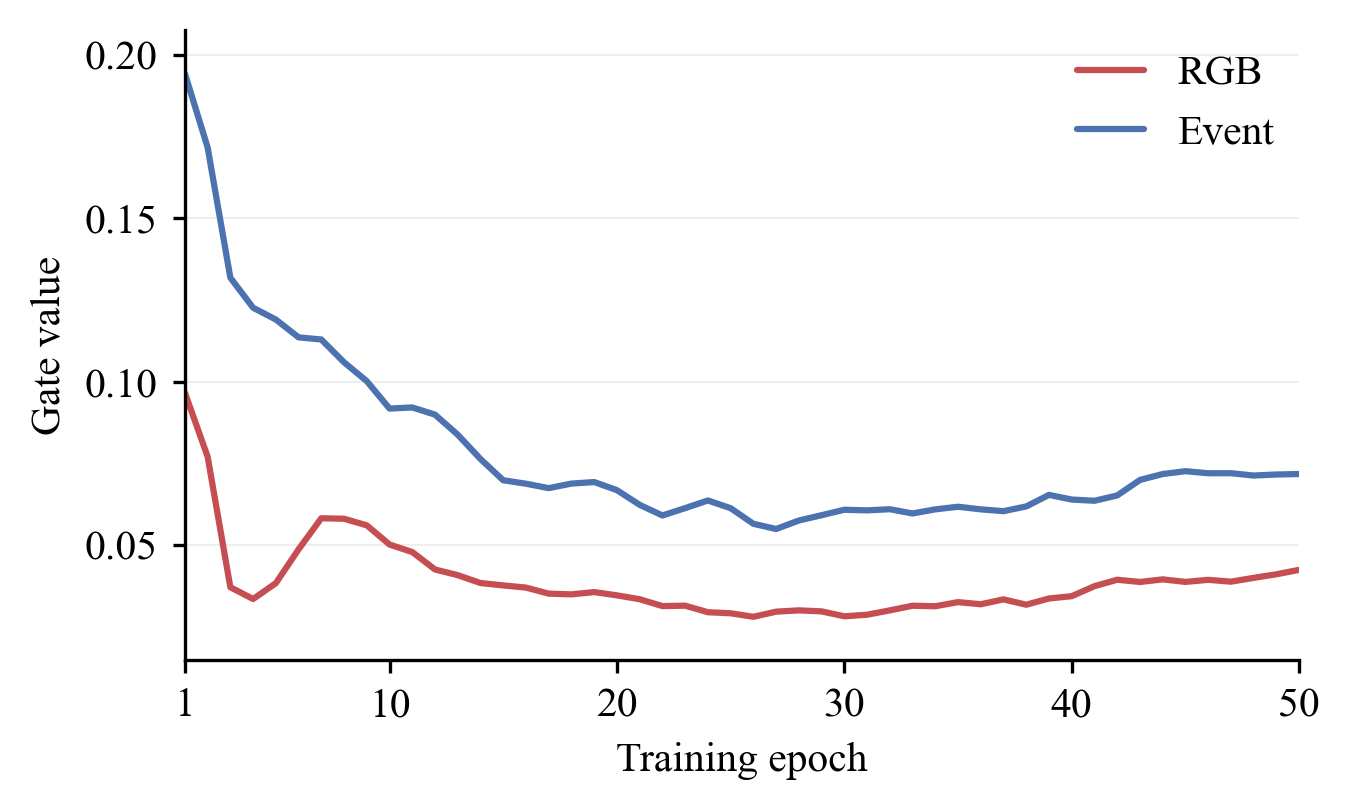}
\caption{Compensation gate dynamics of FCHR during training.}
\label{fig:compensation_gate}
\end{figure}
As shown in Fig.~\ref{fig:compensation_gate}, the Event branch consistently shows higher average compensation gate values at each epoch than the RGB branch, indicating that the Event representation relies more on historical compensation. Since Event responses are triggered by brightness changes, the target may become sparse or even less visible in static or weak-motion scenes, resulting in weak current frame features. The stronger Event compensation gate therefore suggests that FCHR adaptively injects more historical target cues into the Event branch, compensating for insufficient current observations.

\noindent $\bullet$ \textbf{Attention Maps and Response Maps.}
\begin{figure}[tbp]
\centering
\includegraphics[width=\linewidth]{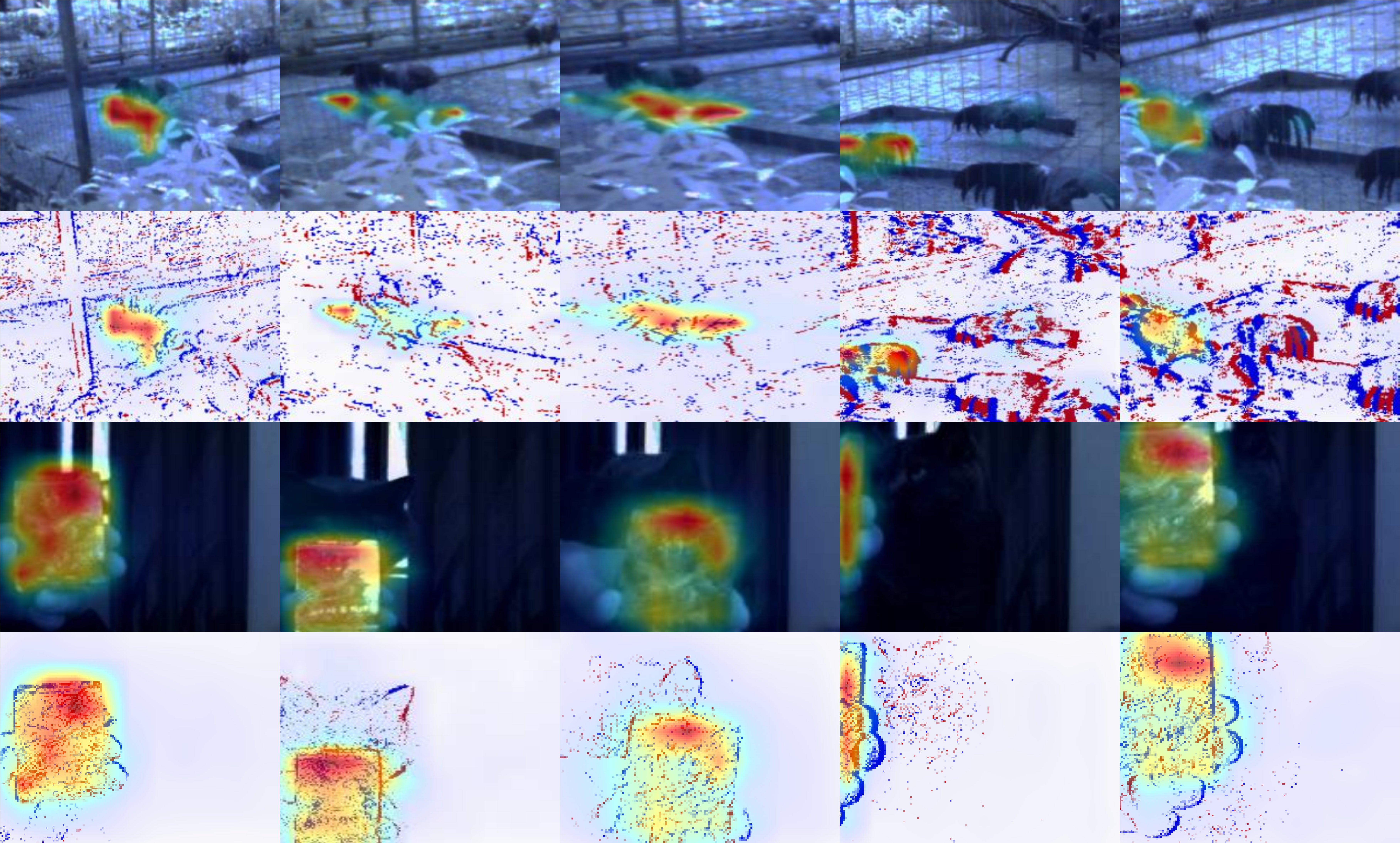}
\caption{Visualization of attention maps generated by APRTrack.}
\label{fig:attention_heatmap}
\end{figure}
\begin{figure}[tbp]
\centering
\includegraphics[width=\linewidth]{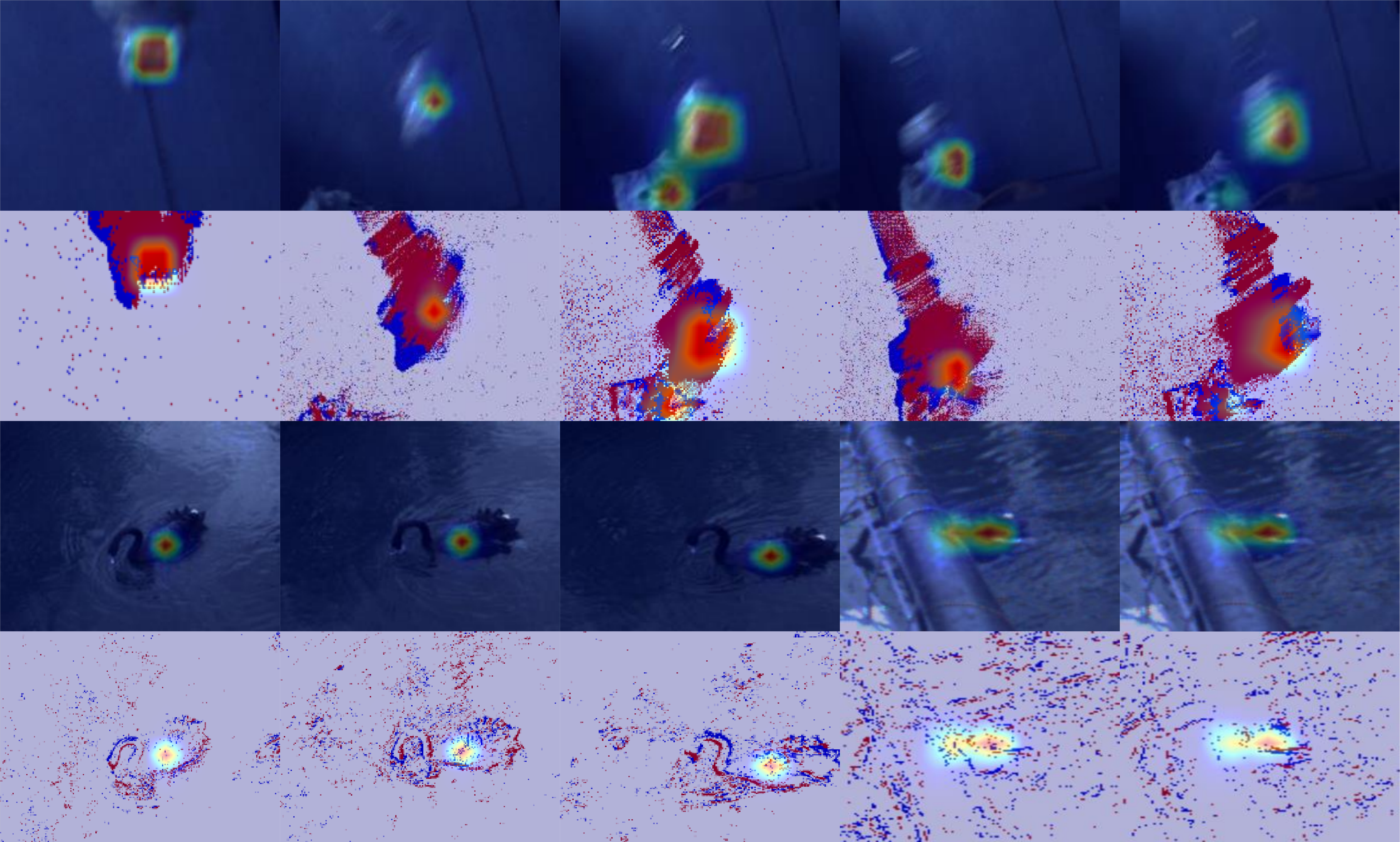}
\caption{Visualization of response maps generated by APRTrack.}
\label{fig:response_heatmap}
\end{figure}
As shown in Fig.~\ref{fig:attention_heatmap}, we visualize the attention activation maps produced by APRTrack in two representative scenarios: similar object interference and target rotation. Regions with warmer colors indicate higher attention weights. The visualization shows that APRTrack can focus on target-related regions despite distractors or appearance changes, rather than being dominated by background regions or similar objects.
Fig.~\ref{fig:response_heatmap} further presents the final response maps generated by the tracker, where high-response regions indicate the predicted target locations. The response maps cover fast motion and partial occlusion scenarios, and APRTrack still produces concentrated responses around the target in the search region. These results suggest that hierarchical perturbation training improves target discrimination under degraded observations, while FCHR helps maintain stable localization when current frame information is incomplete.

\noindent $\bullet$ \textbf{Tracking Results.}
\begin{figure}[tbp]
\centering
\includegraphics[width=\linewidth]{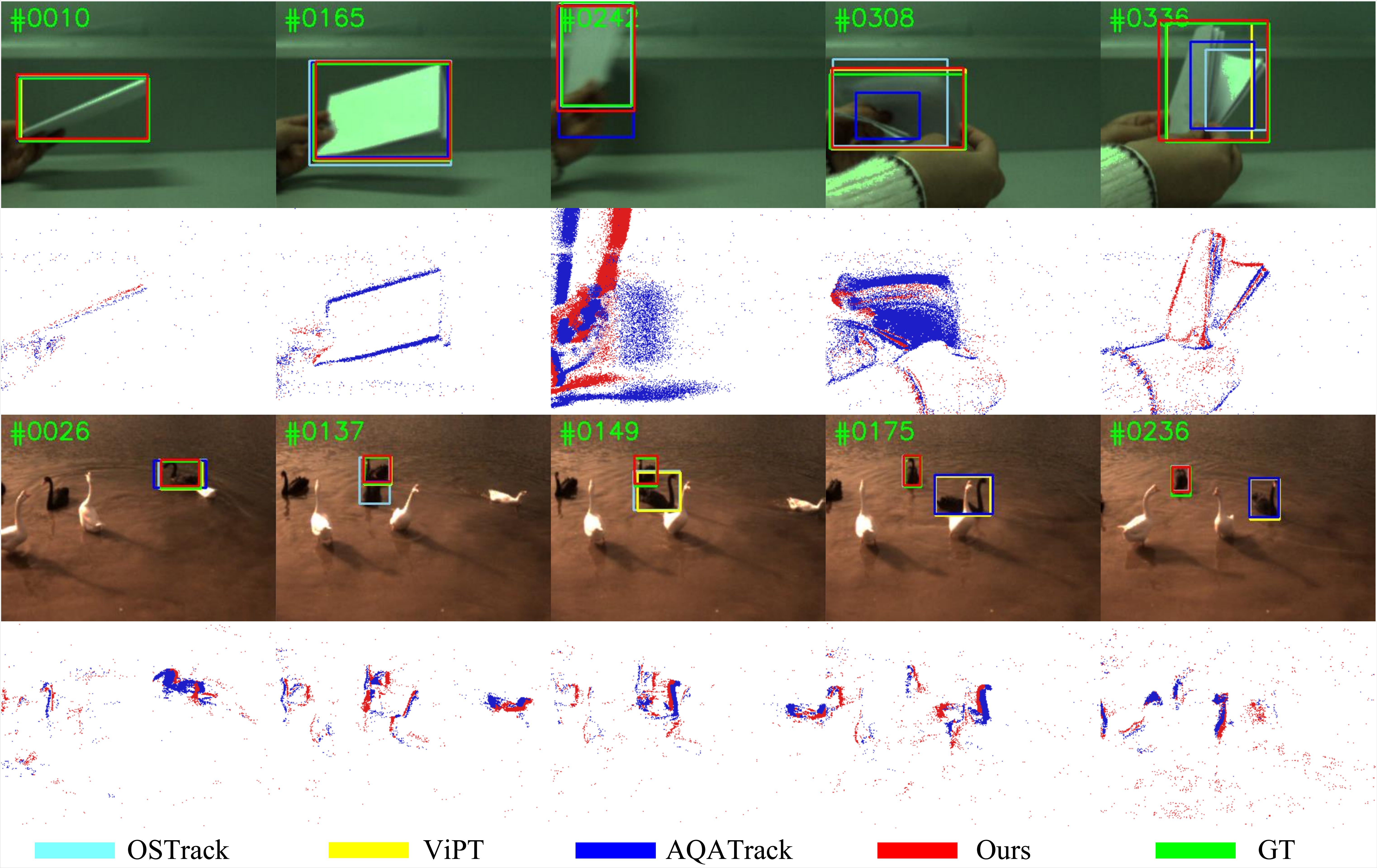}
\caption{Visualization of tracking results produced by APRTrack and other state-of-the-art trackers.}
\label{fig:tracking_res}
\end{figure}
Beyond quantitative analysis, we provide qualitative tracking results for a more intuitive understanding of APRTrack. As shown in Fig.~\ref{fig:tracking_res}, APRTrack is compared with OSTrack~\cite{ye2022joint}, ViPT~\cite{zhu2023visual}, and AQATrack~\cite{xie2024autoregressive} on RGB-Event tracking sequences. Compared with these representative trackers, APRTrack better exploits complementary RGB and Event cues, and produces more stable bounding boxes through missing-robust perturbation training and controlled historical retrieval. These qualitative results further confirm its localization stability in challenging scenes.

\subsection{Limitation Analysis}
Although APRTrack achieves promising performance on multiple RGB-Event tracking benchmarks, it still has limitations. First, our framework mainly focuses on robust representation learning under structured missing conditions, using hierarchical perturbation and historical retrieval to alleviate modality-level degradation and local target corruption. It does not explicitly model long-term state evolution across frames, which may affect performance in sequences with drastic appearance changes or continuous target state transitions. Second, when RGB and Event observations are both severely degraded, or when the historical memory contains unreliable target states, the complementary cues available for recovery may become insufficient. Future work will explore stronger temporal modeling, reliable memory updating, and uncertainty-aware retrieval to improve robustness in long-term and extremely degraded scenarios.

\section{Conclusion}
In this paper, we present APRTrack, an adversarial hierarchical perturbation and retrieval framework for missing-robust RGB-Event tracking. To address structured degradation in real-world scenarios, APRTrack decomposes missing observations into modality-level degradation and local target corruption, and models them with adversarial modality-level perturbation and adversarial spatial-level perturbation, respectively. A hierarchical routing training strategy further decouples different degradation states, avoiding information collapse caused by directly stacking multiple degradations on the same sample. Meanwhile, Footprint-guided Channel-calibrated Hopfield Retrieval incorporates historical target cues in a controlled manner, using query-memory association footprints to calibrate the retrieval space before associative memory access. Experiments on multiple RGB-Event tracking benchmarks show that APRTrack stably integrates modality-robust learning, spatial missing modeling, and reliable historical retrieval, leading to consistent performance improvements. These results validate the effectiveness of unifying structured degradation modeling and controlled memory retrieval within a single framework.

\section*{Acknowledgment} 
This work was supported by the National Natural Science Foundation of China under Grant 62102205 and the Anhui Provincial Natural Science Foundation-Outstanding Youth Project under Grant 2408085Y032.
The authors acknowledge the High-performance Computing Platform of Anhui University for providing computing resources.

\begingroup
\small
\bibliographystyle{IEEEtran}
\bibliography{reference}
\endgroup

\end{document}